\documentclass[letterpaper]{article} 
\usepackage{aaai23}  
\usepackage{times}  
\usepackage{helvet}  
\usepackage{courier}  
\usepackage[hyphens]{url}  
\usepackage{graphicx} 
\usepackage{natbib}  
\usepackage{caption} 
\usepackage{algorithm}
\usepackage{algorithmic}
\usepackage{booktabs}
\usepackage{array}
\newcolumntype{H}{>{\setbox0=\hbox\bgroup}c<{\egroup}@{}}
\usepackage{newfloat}
\usepackage{listings}
\DeclareCaptionStyle{ruled}{labelfont=normalfont,labelsep=colon,strut=off} 
\floatstyle{ruled}
\newfloat{listing}{tb}{lst}{}
\floatname{listing}{Listing}

\usepackage{subfigure}
\usepackage{amsfonts}
\usepackage{multirow}
\usepackage{amsmath}
\usepackage{xcolor}
\usepackage{amssymb}
\usepackage{pifont}
\newcommand{\cmark}{\checkmark} 
\newcommand{\xmark}{\ding{55}} 

\title{Self-Asymmetric Invertible Network for Compression-Aware Image Rescaling}

\author {
    Jinhai Yang\textsuperscript{\rm 1}\equalcontrib,
    Mengxi Guo\textsuperscript{\rm 1}\equalcontrib,
    Shijie Zhao\textsuperscript{\rm 1}\thanks{Corresponding author.},
    Junlin Li\textsuperscript{\rm 2},
    Li Zhang\textsuperscript{\rm 2}
}
\affiliations {
    \textsuperscript{\rm 1} Bytedance Inc., Shenzhen, China\\
    \textsuperscript{\rm 2} Bytedance Inc., San Diego, CA, 92122 USA\\
    \{yangjinhai.01, guomengxi.qoelab, zhaoshijie.0526, lijunlin.li, lizhang.idm\}@bytedance.com
}

\begin{document}

\maketitle

\begin{abstract}
High-resolution (HR) images are usually downscaled to low-resolution (LR) ones for better display and afterward upscaled back to the original size to recover details.
Recent work in image rescaling formulates downscaling and upscaling as a unified task and learns a bijective mapping between HR and LR via invertible networks.
However, in real-world applications (e.g., social media), most images are compressed for transmission.
Lossy compression will lead to irreversible information loss on LR images, hence damaging the inverse upscaling procedure and degrading the reconstruction accuracy.
In this paper, we propose the Self-Asymmetric Invertible Network (SAIN) for compression-aware image rescaling.
To tackle the distribution shift, we first develop an end-to-end asymmetric framework with two separate bijective mappings for high-quality and compressed LR images, respectively.
Then, based on empirical analysis of this framework, we model the distribution of the lost information (including downscaling and compression) using isotropic Gaussian mixtures and propose the Enhanced Invertible Block to derive high-quality/compressed LR images in one forward pass.
Besides, we design a set of losses to regularize the learned LR images and enhance the invertibility.
Extensive experiments demonstrate the consistent improvements of SAIN across various image rescaling datasets in terms of both quantitative and qualitative evaluation under standard image compression formats (i.e., JPEG and WebP).
\textbf{Code is available at \url{https://github.com/yang-jin-hai/SAIN}.}
\end{abstract}

\section{Introduction}\label{sec:intro}

\begin{figure}[t]
\centering
\includegraphics[width=\linewidth]{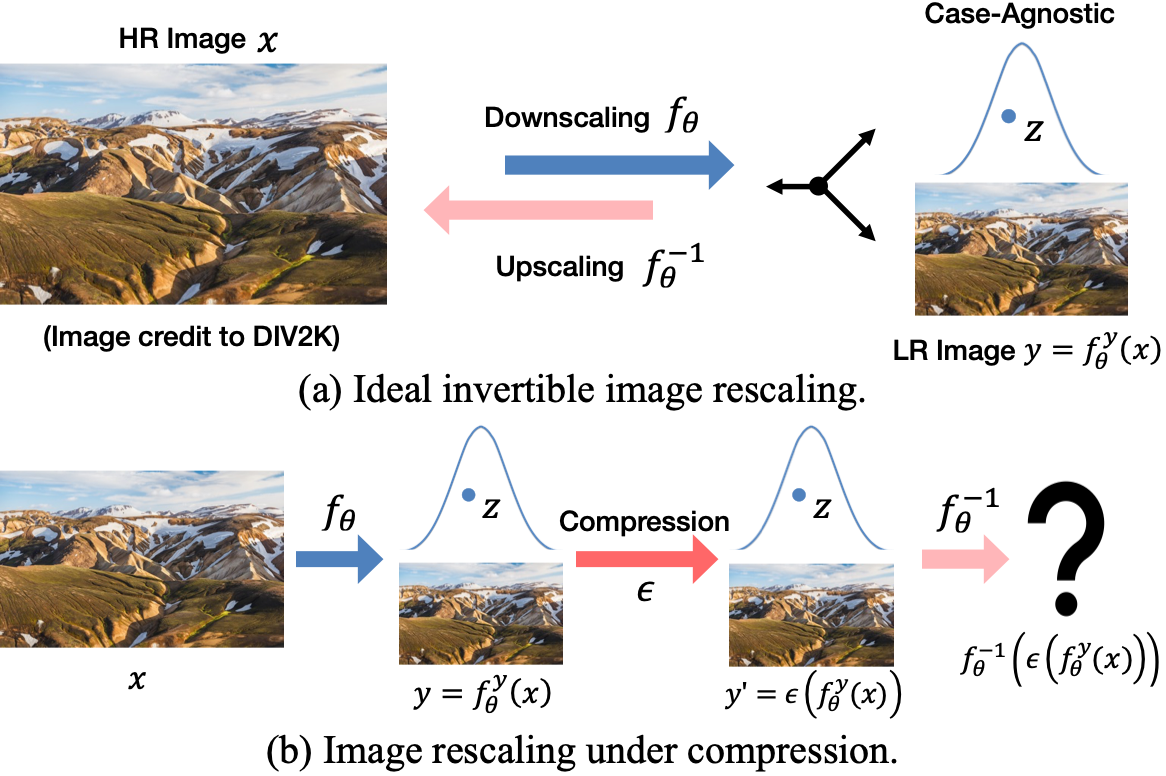}
\caption{Image rescaling without or with compression.}
\label{fig:motivation}
\end{figure}

\begin{figure*}[t]
    \centering
    \includegraphics[width=\linewidth]{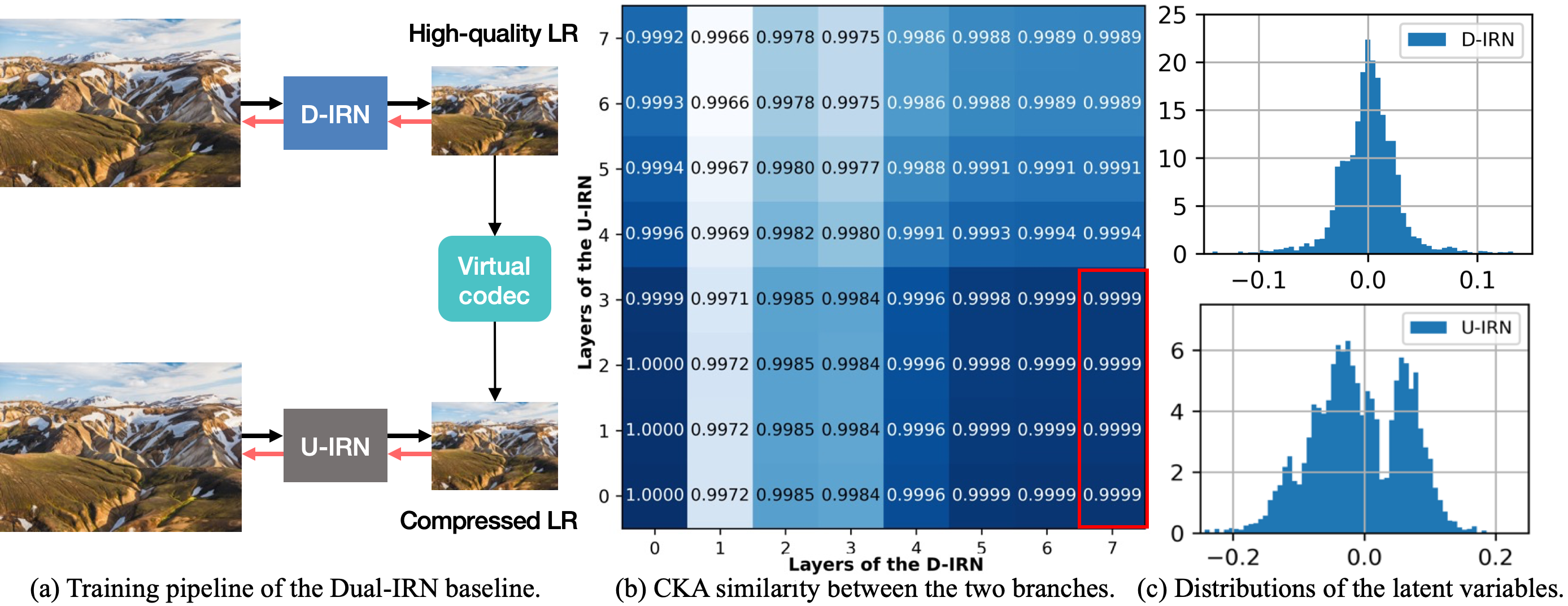}
    \caption{Illustration and empirical analysis of the Dual-IRN baseline under the proposed asymmetric framework.
    The early-layer features of the U-IRN exhibit high similarity to the final output of the D-IRN (i.e., the high-quality LR images).
    Besides, the latent variables of the U-IRN generally have a multi-modal distribution.
    Our SAIN model is inspired by these phenomena.
    }
    \label{fig:dualirn}
\end{figure*}

With advances in computational photography and imaging devices, we are facing increasing amounts of high-resolution (HR) visual content nowadays.
For better display and storage saving, HR images are often downscaled to low-resolution (LR) counterparts with similar visual appearances.
The inverse upscaling is hence indispensable to recover LR images to the original sizes and restore the details. 
Super-resolution algorithms~\cite{dong2015image,dai2019second} have been the prevalent solution to increase image resolution, but they commonly assume the downscaling operator is pre-determined and not learnable.
To enhance the reconstruction quality, recent works~\cite{kim2018task,xiao2020invertible,guo2022invertible} have attempted to jointly optimize the downscaling and upscaling process.
Especially, IRN~\cite{xiao2020invertible} firstly model the conversion between HR and LR image pairs as a bijective mapping with invertible neural networks (INN) to preserve as much information as possible and force the high-frequency split to follow a case-agnostic normal distribution, as shown in Fig.~\ref{fig:motivation}(a).

However, the LR images are usually compressed~\cite{son2021enhanced} to further reduce the bandwidth and storage in realistic scenarios, especially for transmission on social media with massive users.
Worse still, lossy compression (e.g. JPEG and WebP) has become a preference for social networks and websites.
Although standard image compression formats~\cite{wallace1992jpeg,webp} take advantage of human visual characteristics and thus produce visually-similar contents, they still lead to inevitable additional information loss, as shown in Fig.~\ref{fig:motivation}(b).
Due to the bijective nature, the INN-based approaches~\cite{xiao2020invertible,liang2021hierarchical} perform symmetric rescaling and thus are especially sensitive to the distribution shift caused by these compression artifacts.
In this sense, lossy compression can also be utilized as an adversarial attack to poison the upscaling procedure.

In this paper, we tackle compression-aware image rescaling via a Self-Asymmetric Invertible Network (SAIN).
Before delving into the details, we start with empirical analyses of a baseline model Dual-IRN.
To mitigate the influence of compression artifacts, we instantiate the Dual-IRN with an asymmetric framework, establishing two separate bijective mappings, as shown in Fig.~\ref{fig:dualirn}(a).
Under this framework, we can conduct \underline{d}ownscaling with the D-IRN to derive visually-pleasing LR images and then, after compression distortion, use the U-IRN for compression-aware \underline{u}pscaling.
To study the behavioral difference between the two branches, we adopt the CKA metric~\cite{kornblith2019similarity} to measure the representation similarity.
As shown in Fig.~\ref{fig:dualirn}(b), the obtained high-quality LR images (i.e., the final outputs of D-IRN) are highly similar to the anterior-layer features of the U-IRN.
Besides, we plot the histograms of the high-frequency splits in Fig.~\ref{fig:dualirn}(c), which are previously assumed to follow the normal distribution.
Interestingly, this assumption does not hold for the latent variables of the compression-aware U-IRN, which exhibits a multi-modal pattern.

Inspired by the analysis above, we inject our SAIN model with inductive bias.
First, we inherit the asymmetric framework from Dual-IRN, which can upscale from compression-distorted LR images without sacrificing the downscaling quality.
Second, we present a compact network design with Enhanced Invertible Block and decouple the blocks into the downscaling module and the compression simulator, which enables approximating high-quality and compressed LR in one forward pass.
Third, we adopt isotropic Gaussian mixtures to model the joint information loss under the entangled effect of downscaling and compression.

Our main contributions are highlighted as follows:
\begin{itemize}
    \item To our knowledge, this work is the first attempt to study image rescaling under compression distortions. 
    The proposed SAIN model integrates rescaling and compression into one invertible process with decoupled modeling.
    \item We present a self-asymmetric framework with Enhanced Invertible Block and design a series of losses to enhance the reconstruction quality and regularize the LR features.
    \item Both quantitative and qualitative results show that SAIN outperforms state-of-the-art approaches by large margins under standard image codecs (i.e., JPEG and WebP).
\end{itemize}

\section{Related Work}
\noindent\textbf{Invertible Neural Networks.}
Invertible neural networks (INNs) originate from flow-based generative models~\cite{dinh2014nice,dinh2016density}.
With careful mathematical designs, INNs learn a bijective mapping between the source domain and the target domain with guaranteed invertibility.
Normalizing-flow methods~\cite{rezende2015variational,kobyzev2020normalizing} map a high-dimensional distribution (e.g. images) to a simple latent distribution (e.g., Gaussian).
The invertible transformation allows for tractable Jacobian determinant computation, so the posterior probabilities can be explicitly derived and optimized by maximum
likelihood estimation (MLE). 
Recent works have applied INNs to different visual tasks, including super-resolution~\cite{lugmayr2020srflow} and image rescaling~\cite{xiao2020invertible}.

\noindent\textbf{Image Rescaling.}
Super-resolution (SR)~\cite{dong2015image,lim2017enhanced,zhang2018residual} aims to reconstruct the HR image given the pre-downscaled one.
Traditional image downscaling usually adopts low-pass kernels (e.g., Bicubic) for interpolation sub-sampling, which generates over-smoothed LR images due to high-frequency information loss.
Differently, image rescaling~\cite{kim2018task,li2018learning,sun2020learned} jointly optimize downscaling and upscaling as a unified task in an encoder-decoder paradigm.
IRN~\cite{xiao2020invertible} models image rescaling as a bijective transformation with INN to maintain as much information about the HR images.
The residual high-frequency components are embedded into a case-agnostic latent distribution for efficient reconstruction. 
Recently, HCFlow~\cite{liang2021hierarchical} proposes a hierarchical conditional flow to unify image SR and image rescaling tasks in one framework.

However, image downscaling is often accompanied by image compression in applications.
Although flow-based methods perform well in ideal image rescaling, they are vulnerable to lossy compression due to the high reliance on reversibility. 
A subtle interference on the LR images usually causes a considerable performance drop.

\begin{figure*}[ht]
    \centering
    \includegraphics[width=\linewidth]{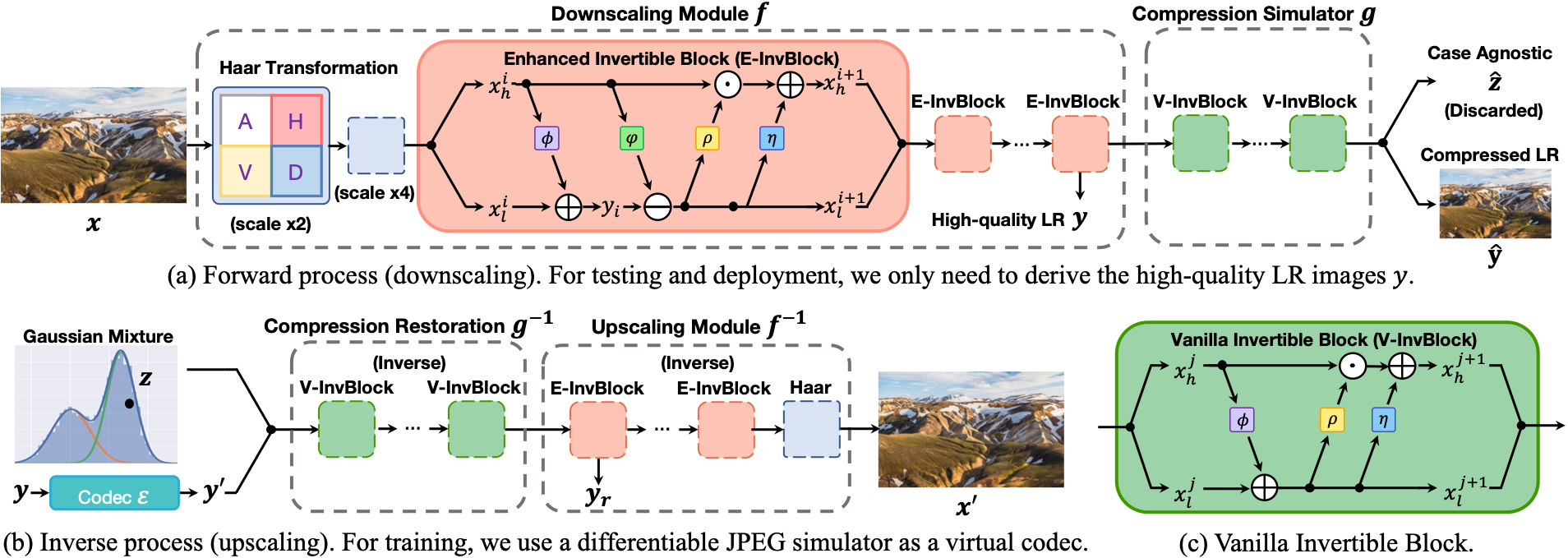}
    \caption{Overview of the proposed Self-Asymmetric Invertible Network (SAIN).
    We decouple the invertible architecture into the downscaling module $f$ to generate visually-appealing LR images and the compression simulator $g$ to mimic standard codecs.
    Thus, due to the reversible nature of InvBlocks, we can inversely restore the compression distortion via $g^{-1}$  from LR images that are compressed by real codecs (albeit not perfectly), and then upscale to original size via $f^{-1}$.
    The $\exp(\cdot)$ of $\rho$ is omitted.
    }
    \label{fig:method}
\end{figure*}

\section{Methodology}
\subsection{Preliminaries}
Normalizing flow models~\cite{kobyzev2020normalizing} usually propagate high-dimensional distribution (e.g. images) through invertible transformation to enforce a simple specified distribution.
The Jacobian determinant of such invertible transformation is easy to compute so that we can inversely capture the explicit distribution of input data and minimize the negative log-likelihood in the source domain.

In the image rescaling task, however, the target domain is the LR images, whose distribution is implicit.
Therefore, IRN~\cite{xiao2020invertible} proposes to split HR images $x$ into low-frequency and high-frequency components $[x_l,x_h]$ and learn the invertible mapping $[x_l,x_h] \leftrightarrow[y, a]$, where $y$ is the desired LR image and $a\sim \mathcal{N}(0,1)$ is case-agnostic.

In this work, we establish an asymmetric framework to enhance the robustness against image compression for the rescaling task.
In the forward approximation pass, we not only model the downscaling process by $f$ but also simulate the compressor by $g$.
The downscaling module $f$ alone can derive the visually-pleasing LR $y$, while the function composition $f\circ g$ maps $[x_l, x_h]\rightarrow[\hat{y},\hat{z}]$, where $\hat{y}$ is the simulated compressed LR.
When $y$ suffers from compression distortion $\varepsilon$ and results in $y^{\prime}$, the reverse restoration pass is used to reconstruct the HR contents by $[y^{\prime},z]\rightarrow x^\prime$.
The forward and inverse pass are asymmetric but share one invertible structure.
The ultimate goal is to let $x^\prime$ approach the true HR $x$.

\subsection{Self-Asymmetric Invertible Network}
The difficulty of compression-robust image rescaling lies in designing a model that performs both high-quality downscaling and compression-distorted upscaling.
Since we approximate the downscaling and the compression process in one forward pass, and they essentially share large proportions of computations, it is important to avoid confusion in the low-frequency split.
In this work, we assume that the information loss caused by compression is conditional on the high-frequency components, and thus devise the Enhanced Invertible Block (E-InvBlock).
During upscaling, the latent variable $z$ is sampled from a learnable Gaussian mixture to help recover details from perturbed LR images.
The overall framework is illustrated in Fig.~\ref{fig:method}.

\paragraph{Haar Transformation.}
We follow existing works~\cite{xiao2020invertible,liang2021hierarchical} to split the input image via the Haar transformation.
This channel splitting is crucial for the construction of invertible modules~\cite{kingma2018glow,ho2019flow++}.
Haar transformation decomposes an image into a low-pass approximation, the horizontal, vertical, and diagonal high-frequency coefficients~\cite{lienhart2002extended}.
The low-frequency (LF) approximation and the high-frequency (HF) components represent the input image as $[x_l,x_h]$.
For a scale larger than 2, we adopt successive Haar transformations to split the channels at the first.

\paragraph{Vanilla Invertible Block.}
The Vanilla Invertible Block (V-InvBlock) inherits the design in IRN~\cite{xiao2020invertible}.
It is a rearrangement of existing coupling layers~\cite{dinh2014nice,dinh2016density} to fit the image rescaling task.
For the $j$-th layer,

\begin{align}
x_{l}^{j+1}&=x_{l}^{j}+\phi\left(x_{h}^{j}\right)~, \\
x_{h}^{j+1}&=x_{h}^{j} \odot \exp \left(\rho\left(x_{l}^{j+1}\right)\right)+\eta\left(x_{l}^{j+1}\right)~,
\end{align}
where $\odot$ denotes the Hadamard product and in practice, we use a centered Sigmoid function for numerical stability after the exponentiation.
The inverse step is easily obtained by
\begin{align}
x_{h}^{j}&=\left(x_{h}^{j+1}-\eta\left(x_{l}^{j+1}\right)\right) \odot \exp \left(-\rho\left(x_{l}^{j+1}\right)\right)~, \\
x_{l}^{j}&=x_{l}^{j+1}-\phi\left(x_{h}^{j}\right)~.
\end{align}

\paragraph{Enhanced Invertible Block.}
In V-InvBlock, the LF part is polished by a shortcut connection on the HF branch.
Since $f$ is cascaded to $g$, simply repeating the V-InvBlock to construct $f$ would cause ambiguity in the LF branch.
Therefore, we augment the LF branch and thus make the modeling of the high-quality LR $y$ and the emulated compressed LR $\hat{y}$ separable to some extent. 
Generally, we let $x_l$ help stimulate $\hat{y}$, and let the intermediate representation $y_i$ undertake the polishing for $y$.
Since there is no information loss inside the block, we assume the compression distortion can be recovered from the HF components.
Formally,
\begin{align}
y_{i}&=x_{l}^{i}+\phi\left(x_{h}^{i}\right)~, \\
x_{l}^{i+1}&=y_{i}-\varphi\left(x_{h}^{i}\right)~.
\end{align}

It only brings a slight increase in computational overhead but significantly increases the model capacity.
Note that $\phi(\cdot)$, $\varphi(\cdot)$, $\eta(\cdot)$, and $\rho(\cdot)$ can be arbitrary functions.

\paragraph{Isotropic Gaussian Mixture.}
The case-specific information is expected to be completely embedded into the downscaled image, since preserving the HF components is impractical.
IRN~\cite{xiao2020invertible} forces the case-agnostic HF components to follow $N(0,I)$ and sample from the same distribution for inverse upscaling.
However, due to the mismatch between real compression and simulated compression, the distribution of the forwarded latent $\hat{z}$ and the underlying upscaling-optimal latent $z$ are arguably not identical.
Besides, as shown in Fig.~\ref{fig:dualirn}(c), the latent distribution of the compression-aware branch presents a multimodal pattern.
Therefore, rather than explicitly modeling the distribution of $\hat{z}$, we choose to optimize a learnable Gaussian mixture to sample for upscaling from compressed LR images.
For simplicity, we assume the Gaussian mixture is isotopic~\cite{amendola2020maximum} and all dimensions of $z$ follow the same univariate marginal distribution. For any $z_o\in z$: 
\begin{equation}
p(z_o)=\sum_{k=1}^{K} \pi_{k} \mathcal{N}\left(z_o\mid\mu_{k}, \sigma_{k}\right)~,
\end{equation}
where the mixture weights $\pi_k$, means $\mu_k$, and variances $\sigma_k$ are learned globally.
Since the sampling operation is non-differentiable, enabling end-to-end optimization of the parameters $\{\pi_k,\mu_k,\sigma_k\}$ is non-trivial.
We decompose the sampling from $p(z_o)$ into two independent steps: (1) discrete sampling $k \sim \text{Categorical} (\pi)$ to select a component; (2) sample $z_o$ from the parameterized $\mathcal{N}(\mu_k,\sigma_k)$.
In this way, we use Gumbel-Softmax~\cite{jang2017categorical} to approximate the first step and use the reparameterization trick~\cite{kingma2013auto} for the second step, and thus estimate the gradient for backpropagation.

\paragraph{Compression and Quantization.}
To jointly optimize the upscaling and the downscaling steps under compression artifacts, we employ a differentiable JPEG simulator~\cite{xing2021invertible} to serve as a virtual codec $\varepsilon$.
It performs discrete cosine transform on each 8$\times$8 block of an image and simulates the rounding function with the Fourier series.
For the LR and HR images, we use the Straight-Through Estimator~\cite{bengio2013estimating} to calculate the gradients of the quantization module.
Moreover, we incorporate real compression distortion $\epsilon$ to provide guidance for network optimization, which extends our model to be also suitable for other image compression formats besides JPEG.

\subsection{Training Objectives}
The downscaling module, the compression simulator, and the inverse restoration procedure are jointly optimized.
The overall loss function is a linear combination of the final reconstruction loss and a set of LR guidance to produce visually-attractive LR images and meanwhile enhance the invertibility:
\begin{equation}
\mathcal{L}=\lambda_{1} \mathcal{L}_{{rec }}+\lambda_{2} \mathcal{L}_{fit}+\lambda_{3} \mathcal{L}^{\prime}_{ {fit }}+\lambda_{4} \mathcal{L}_{ {reg }}+\lambda_{5} \mathcal{L}_{ {rel }}~.
\end{equation}

\paragraph{HR Reconstruction.}
Despite the information loss caused by the downscaling and the compression, we expect that given a model-downscaled LR $y$, the counterpart HR image $x$ can be restored by our model using a random sample of $z$ from the learned distribution $p(z_o)$:
\begin{equation}
\mathcal{L}_{rec}=\mathcal{L}_{hr}\left(f^{-1}\left(g^{-1}\left(\left[y,z\right]\right)\right), x\right)~.
\end{equation}

\paragraph{LR Guidance.}
First, the model-downscaled LR images should be visually-meaningful.
We follow existing image rescaling works~\cite{kim2018task,xiao2020invertible} to drive the LR images $y$ to resemble Bicubic interpolated images as a guidance of the downscaling module $f$:
\begin{equation}
\mathcal{L}_{fit}=\mathcal{L}_{lr}\left(\operatorname{Bicubic}\left(x\right),y\right)~.
\end{equation}
Second, to better simulate the compression distortions and the inverse restoration, we encourage the model-distorted LR image $\hat{y}$ to approximate the compressed version of the Bicubic downscaled LR image which undergoes the distortion $\epsilon$ of a real image compression process:
\begin{equation}
\mathcal{L}_{fit}^{\prime}=\mathcal{L}_{lr}\left(\epsilon\left(\operatorname{Bicubic\left(x\right)}\right),\hat{y}\right)~.
\end{equation}

Third, we regularize the similarity between the model-downscaled LR image $y$ and the inversely restored LR image $y_r$ to enhance reversibility: $\mathcal{L}_{reg}=\mathcal{L}_{lr}\left(y, y_r\right)$.

Finally, we further facilitate the compression simulation by enforcing the relation between $y$ and $\hat{y}$: $\mathcal{L}_{rel}=\mathcal{L}_{lr}\left(\epsilon\left(y\right),\hat{y}\right)$.
Note that $\epsilon$ can be any image compressor to make our model robust against other compression formats.

\section{Experiments}
\begin{table*}[t]
    \centering
    \small
\begin{tabular}{lccccccccccccccccccc}\toprule
 Downscaling \& Upscaling & Scale  & JPEG QF=30& JPEG QF=50 & JPEG QF=70 & JPEG QF=80& JPEG QF=90  \\\midrule
 Bicubic \& Bicubic & $\times2$ & 29.38 / 0.8081&
30.19 / 0.8339&
30.91 / 0.8560&
31.38 / 0.8703&
31.96 / 0.8878\\
Bicubic \& SRCNN~\cite{dong2015image} & $\times 2$  &28.01 / 0.7872&
28.69 / 0.8154&
29.43 / 0.8419&
30.01 / 0.8610&
30.88 / 0.8878  \\
Bicubic \& EDSR~\cite{lim2017enhanced} & $\times 2$  & 28.92 / 0.7947&
29.93 / 0.8257&
31.01 / 0.8546&
31.91 / 0.8753&
33.44 / 0.9052 \\
 Bicubic \& RDN~\cite{zhang2018residual} & $\times 2$  & 28.95 / 0.7954&
29.96 / 0.8265&
31.02 / 0.8549&
31.91 / 0.8752&
33.41 / 0.9046\\
 Bicubic \& RCAN~\cite{zhang2018image} & $\times 2$ & 28.84 / 0.7932&
29.84 / 0.8245&
30.94 / 0.8538&
31.87 / 0.8749&
33.44 / 0.9052\\
 CAR \& EDSR~\cite{sun2020learned}& $\times 2$ & 27.83 / 0.7602 &
28.66 / 0.7903 &
29.44 / 0.8165 &
30.07 / 0.8347 &
31.31 / 0.8648\\
 IRN~\cite{xiao2020invertible} & $\times 2$ & 29.24 / 0.8051&
30.20 / 0.8342&
31.14 / 0.8604&
31.86 / 0.8783&
32.91 / 0.9023 \\
 SAIN (Ours)& $\times 2$ & \textbf{31.47 / 0.8747}&
\textbf{33.17 / 0.9082}&
\textbf{34.73 / 0.9296}&
\textbf{35.46 / 0.9374}&
\textbf{35.96 / 0.9419}\\\midrule
 Bicubic \& Bicubic & $\times 4$& 26.27 / 0.6945&
26.81 / 0.7140&
27.28 / 0.7326&
27.57 / 0.7456&
27.90 / 0.7618  \\
 Bicubic \& SRCNN~\cite{dong2015image} & $\times 4$ &25.49 / 0.6819&
25.91 / 0.7012&
26.30 / 0.7206&
26.55 / 0.7344&
26.84 / 0.7521\\
 Bicubic \& EDSR~\cite{lim2017enhanced} & $\times 4$ &25.87 / 0.6793&
26.57 / 0.7052&
27.31 / 0.7329&
27.92 / 0.7550&
28.88 / 0.7889\\
 Bicubic \& RDN~\cite{zhang2018residual} & $\times 4$ & 25.92 / 0.6819&
26.61 / 0.7075&
27.33 / 0.7343&
27.92 / 0.7556&
28.84 / 0.7884\\
 Bicubic \& RCAN~\cite{zhang2018image} & $\times 4$ & 25.77 / 0.6772&
26.45 / 0.7031&
27.21 / 0.7311&
27.83 / 0.7537&
28.82 / 0.7884\\
 Bicubic \& RRDB~\cite{wang2018esrgan} & $\times 4$ & 25.87 / 0.6803&
26.58 / 0.7063&
27.36 / 0.7343&
27.99 / 0.7568&
28.98 / 0.7915 \\
 CAR \& EDSR~\cite{sun2020learned}&  $\times 4$ & 25.25 / 0.6610&
25.76 / 0.6827&
26.22 / 0.7037&
26.69 / 0.7214&
27.91 / 0.7604\\
 IRN~\cite{xiao2020invertible} & $\times 4$ &25.98 / 0.6867&
26.62 / 0.7096&
27.24 / 0.7328&
27.72 / 0.7508&
28.42 / 0.7777 \\
 HCFlow~\cite{liang2021hierarchical}& $\times 4$&25.89 / 0.6838&
26.38 / 0.7029&
26.79 / 0.7204&
27.05 / 0.7328&
27.41 / 0.7485\\
SAIN (Ours)& $\times 4$ & \textbf{27.90 / 0.7745}&
\textbf{29.05 / 0.8088}&
\textbf{29.83 / 0.8272}&
\textbf{30.13 / 0.8331}&
\textbf{30.31 / 0.8367}\\\bottomrule
\end{tabular}
    \caption{Quantitative results (PSNR / SSIM) of image rescaling on DIV2K under distortion at different JPEG QFs. }
    \label{tab:DIV2K}
\end{table*}
\begin{figure*}[t]
    \centering
    \includegraphics[width=\linewidth]{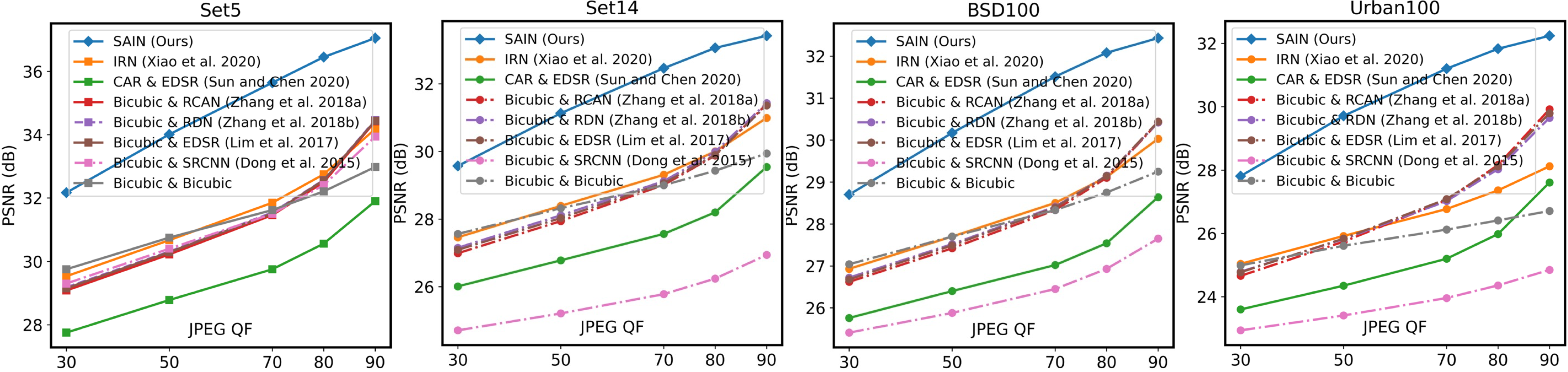}
    \caption{Cross-dataset evaluation of image rescaling ($\times2$) over standard benchmarks: Set5, Set14, BSD100, and Urban100. }
    \label{fig:benckmark}
\end{figure*}

\subsection{Experimental Setup}
\paragraph{Datasets and Settings.}
We adopt the 800 HR images from the widely-acknowledged DIV2K training set~\cite{agustsson2017ntire} to train our model.
Apart from the DIV2K validation set, we also evaluate our model on 4 standard benchmarks: Set5~\cite{bevilacqua2012low}, Set14~\cite{zeyde2010single}, BSD100~\cite{martin2001database}, and Urban100~\cite{huang2015single}.
Following the convention in image rescaling~\cite{xiao2020invertible,liang2021hierarchical}, the evaluation metrics are Peak Signal-to-Noise Ratio (PSNR) and SSIM~\cite{wang2004image} on the Y channel of the YCbCr color space.

\paragraph{Implementation Details.}
For $\times2$ and $\times4$ image rescaling, we use a total of 8 and 16 InvBlocks in total, and the downscaling module $f$ has 5 and 10 E-InvBlocks, respectively.
The transformation functions $\phi(\cdot)$, $\varphi(\cdot)$, $\eta(\cdot)$, and $\rho(\cdot)$ are implemented with Dense Block~\cite{wang2018esrgan,xiao2020invertible}.
The input images are cropped to 128$\times$128 and augmented via random horizontal and vertical flips.
We adopt Adam optimizer~\cite{kingma2014adam} with $\beta_1=0.9$ and $\beta_2=0.999$, and set the mini-batch size to 16.
The model is trained for 500$k$ iterations. 
The learning rate is initialized as $2\times10^{-4}$ and reduced by half every $100k$ iterations.
We use $\mathcal{L}_1$ pixel loss as the LR guidance loss $\mathcal{L}_{lr}$ and $\mathcal{L}_2$ pixel loss as the HR reconstruction loss.
To balance the losses in LR and HR spaces, we use $\lambda_1=1$ and $\lambda_2=\lambda_3=\lambda_4=\lambda_5=\frac{1}{4}$.
The compression quality factor (QF) is empirically fixed at 75 during training. 
The Gaussian mixture for upscaling has $K=5$ components.

\subsection{Evaluation under JPEG Distortion}
JPEG~\cite{wallace1992jpeg} is the most widely-used lossy image compression method in consumer electronic products.
As various compression QFs may be used in applications, we expect SAIN to be a one-size-fits-all model.
That is, after compression-aware training under a specific format, it can handle compression distortions at different QFs.

\begin{figure*}[!ht]
    \centering
    \includegraphics[width=\linewidth]{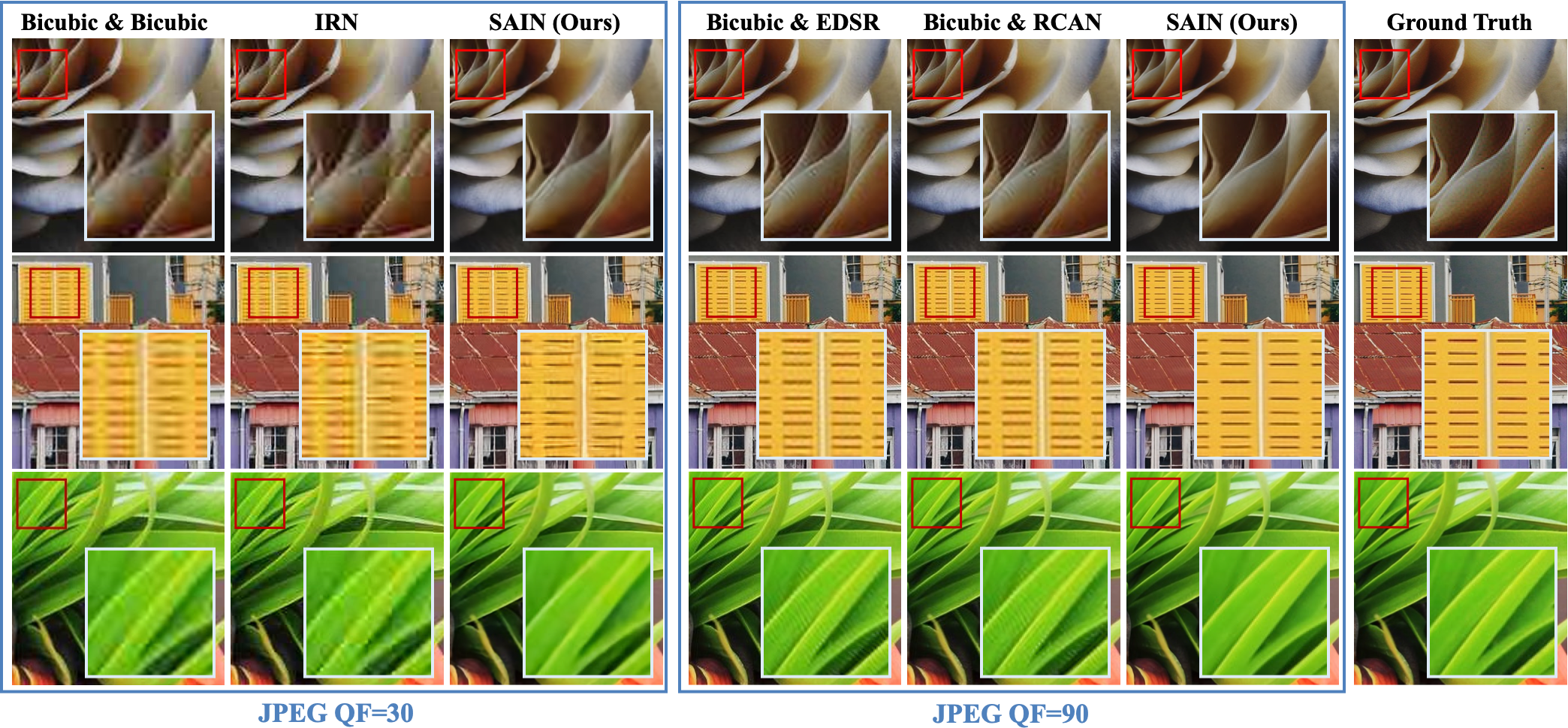}
    \caption{Qualitative results of image rescaling ($\times$2) on DIV2K under distortion at different JPEG QFs. Due to space limitation, we only visualize results of the top three methods at each QF. More results can be found in the supplementary material.}
    \label{fig:visual}
\end{figure*}

\paragraph{Quantitative Evaluation.}
We compare our model with three kinds of methods: (1) Bicubic downscaling \& super-resolution~\cite{dong2015image,lim2017enhanced,zhang2018image,zhang2018residual}; (2) jointly-optimized downscaling and upscaling~\cite{sun2020learned}; (3) flow-based invertible rescaling models~\cite{xiao2020invertible,liang2021hierarchical}.
We reproduce all the compared methods and get similar performance as claimed in their original papers when tested without compression distortions. 
Then, for each approach, we apply JPEG distortion to the downscaled LR images at different QFs (i.e., 30, 50, 70, 80, 90), and evaluate the construction quality of the upscaled HR images.

As is evident in Tab.~\ref{tab:DIV2K}, the proposed SAIN outperforms existing methods by a large margin at both scales and across all testing QFs.
The reconstruction PSNR of SAIN significantly surpasses the second best results by 1.33-3.59 dB.
The performances of previous state-of-the-arts drop severely as JPEG QF descents and are even worse than naïve Bicubic resampling at lower QF and larger scale. 
HCFlow~\cite{liang2021hierarchical} suffers even more from the compression artifacts than IRN~\cite{xiao2020invertible} since it assumes the HF components are conditional on the LF part.

\paragraph{Cross-Dataset Validation.}
In addition to the DIV2K validation set, we further verify our methods on 4 standard benchmarks (Set5, Set14, BSD100, Urban100) to investigate the cross-dataset performance.
Similarly, we test all the compared models at different QFs for the $\times2$ image rescaling task, and plot the corresponding PSNR in Fig.~\ref{fig:benckmark}.

From these curves, we can clearly observe that the proposed approach SAIN achieves substantial improvements over the state-of-the-arts.
SRCNN~\cite{dong2015image} adopts a very shallow convolutional network to increase image resolution and thus is quite fragile to the compression artifacts.
CAR \& EDSR~\cite{sun2020learned} learns an upscaling-optimal downscaling network in an encoder-decoder framework that is specific to EDSR~\cite{lim2017enhanced}, so when the downscaled LR images are distorted by compression, it also fails at restoring high-quality HR images.
Interestingly, there exists a trade-off between the tolerance to high QFs and the low QFs for the other methods.
Those who perform better at higher QFs seem to be inferior at lower QFs.
Differently, our model consistently performs better due to the carefully designed network architecture and training objectives. 

\paragraph{Qualitative Evaluation.}
Fig.~\ref{fig:visual} shows the visual details of the top HR reconstruction results at QF=30 and QF=90.
The chroma subsampling and quantization in JPEG encoding lead to inevitable information loss, hence the compared methods tend to produce blurred and noisy results.
Apparently, SAIN can better restore image details and still generate sharp edges at a low JPEG QF of 30, which is attributed to the proposed compression-aware invertible structure.

Besides, although our model implicitly embeds all HF information into the LR images, they remain similar appearances to the Bicubic interpolated ground-truth.
Some downscaled LR results are shown in the supplementary material. 

\subsection{Ablation Study}\label{sec:ablation}
\paragraph{Ablation Study on Training Strategy.}
To prove the effectiveness and efficiency of our model, we conduct an ablation study on the training strategy.
We design a set of alternatives with three training strategies: 
(1) Vanilla: training without compression-awareness.
(2) Fine-tuning: the downscaling process is pre-defined while the upscaling process is finetuned with compression-distorted LR images; 
(3) Ours: the downscaling and upscaling are jointly optimized with the proposed asymmetric framework (recall Fig~\ref{fig:dualirn}).

Tab.~\ref{tab:alters} lists the quantitative results of $\times2$ image rescaling evaluated at JPEG QF=75.
The models following the proposed asymmetric training framework are evidently superior to the two-stage finetuning methods.
Compared with IRN~\cite{xiao2020invertible}, our model only increases 0.36M parameters, but significantly boosts the performance by 3.65~dB.
Compared with Dual-IRN, SAIN reduces $\sim40\%$ parameters and still achieves a 0.41 dB gain.

\begin{table}[t]
    \centering
    \small
    \begin{tabular}{l|cc|c|c|cccccc}\toprule
     Strategy &CA & AF & Method & Param & PSNR \\\midrule
    Vanilla &\xmark&\xmark&IRN&$1.66\mathrm{M}$&31.45 \\\cmidrule{1-6}
    \multirow{3}{.15\linewidth}{Fine-Tuning}&\checkmark&\xmark&IRN &$1.66\mathrm{M}$&32.70\\
    &\checkmark&\xmark&Bicubic\&EDSR&$40.7\mathrm{M}$&32.97\\
    &\checkmark&\xmark&IRN\&EDSR &$42.4\mathrm{M}$&32.92\\\cmidrule{1-6}
    \multirow{3}{.15\linewidth}{Ours}&\checkmark&\checkmark&IRN\&EDSR &$42.4\mathrm{M}$&\textbf{34.41} \\
    &\checkmark&\checkmark&Dual-IRN &$3.34\mathrm{M}$&\textbf{34.69}\\
    &\checkmark&\checkmark&SAIN (Ours)&$2.02\mathrm{M}$&\textbf{35.10} \\\bottomrule
    \end{tabular}  
    \caption{Ablation study on training strategy. 
    ``CA'': Training with \underline{c}ompression-\underline{a}wareness. 
    ``AF'': Training with the proposed \underline{a}symmetric \underline{f}ramework. 
    Refer to Sec.~\ref{sec:ablation} for details.}
    \label{tab:alters}
\end{table}

\paragraph{Ablation Study on Training Objective.}
In this part, we investigate the effect of the introduction of the Gaussian mixture model (GMM) and the additional training objectives. 
As presented in Tab.~\ref{tab:ablations}, all of them play a positive role in the final performance.
The $\mathcal{L}_{fit}^{\prime}$ matters most since it guides the sub-network $g$ on how to simulate real compression. 
Thanks to the invertibility, it simultaneously learns the $g^{-1}$ for restoration from real compression artifacts. 
Different from IRN~\cite{xiao2020invertible} that captures the HF components with a case-agnostic distribution $\mathcal{N}(0,1)$, we utilize a learnable GMM to excavate a universal knowledge about the HF component.
Although this distribution is learned on DIV2K, we can see that it also improves the performance on other datasets (e.g., Set5).

\begin{figure}[t]
    \centering
    \includegraphics[width=\linewidth]{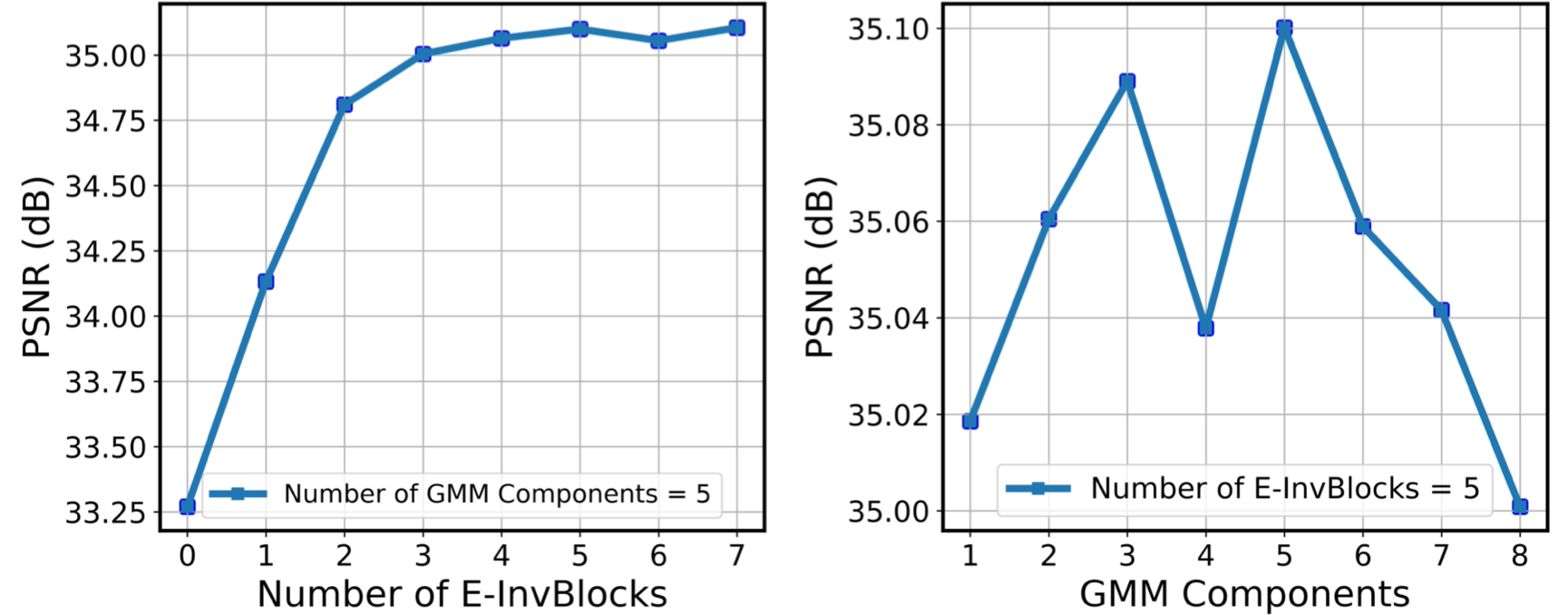}
    \caption{Effect of the number of E-InvBlocks and GMM components. Evaluated at JPEG QF=75 on DIV2K ($\times2$).}
    \label{fig:ablations}
\end{figure}

\section{Further Analysis}
\subsection{Hyper-Parameter Selection}
We mainly investigate the influence of different settings of the hyper-parameters related to the model structure.
We fixed the total number of blocks as 8 (for $\times2$ task) and search for the best value of the number of E-InvBlocks.
It actually searches for a complexity balance between the downscaling module $f$ and the compression simulator $g$.
From Fig.~\ref{fig:ablations}, we observe that the performance saturates after this value reaches 5.
Since adding E-InvBlocks would increase the number of parameters, we use 5 and 10 E-InvBlocks for $\times2$ and $\times4$ experiments, respectively. 
Besides, we can find that the best value of GMM components $K$ is 5.
Too large or too small values both degrade the performance.
An analysis of the effect of the training QF is in the supplementary.

\subsection{Evaluation under WebP Distortion}
WebP~\cite{webp} is a modern compression format that is widely used for images on the Internet. 
Therefore, we further validate the tolerance against WebP of the proposed model.
To make the optimization end-to-end, we still use the differentiable JPEG simulator as the virtual codec $\varepsilon$ to approximate the gradient but set WebP as the real compression $\epsilon$ to guide the behavior of the invertible compression simulator $g$.
We test the image rescaling ($\times2$) performance at the WebP QF of 30 and 90.
Tab.~\ref{tab:webpx2} again shows the robustness against real compression artifacts of our model, which demonstrates the potential to extend to other image compression methods.

\begin{table}[t]
    \centering
    \begin{tabular}{c|ccc|ccc}\toprule
         \multirow{2}{*}{
         \begin{tabular}{c}
              $\mathcal{N}(0, 1)$ \\ $\rightarrow$GMM\\
         \end{tabular}
         } 
         & \multirow{2}{*}{$\mathcal{L}_{rel}$}&\multirow{2}{*}{$\mathcal{L}_{fit}^{\prime}$}&\multirow{2}{*}{$\mathcal{L}_{reg}$} &\multirow{2}{*}{Set5} &\multirow{2}{*}{DIV2K}\\
         &&&&\\\midrule
         &\cmark &\cmark&\cmark&35.98 & 35.01\\
         \cmark& &\cmark&\cmark&36.02 & 35.02\\
         \cmark&\cmark &&\cmark&35.85 & 34.95\\
         \cmark&\cmark &\cmark&&35.95 & 34.96\\
         \cmark&\cmark &\cmark&\cmark&\textbf{36.04} & \textbf{35.10} \\\bottomrule
    \end{tabular}
    \caption{Ablation study of PSNR under JPEG QF=75.}
    \label{tab:ablations}
\end{table}

\begin{table}[t]
    \centering
    \small
\begin{tabular}{lccccccccccccccccccc}\toprule
 Downscaling \& Upscaling & QF  & PSNR / SSIM \\\midrule
 Bicubic \& Bicubic & 90 & 32.02 / 0.8922\\
Bicubic \& SRCNN~\cite{dong2015image} &90  & 31.29 / 0.9014 \\
Bicubic \& EDSR~\cite{lim2017enhanced} &90   & 34.32 / 0.9220 \\
 Bicubic \& RDN~\cite{zhang2018residual} &90  & 34.26 / 0.9212\\
 Bicubic \& RCAN~\cite{zhang2018image} &90  & 34.34 / 0.9222\\
 CAR \& EDSR~\cite{sun2020learned}&90 & 32.58 / 0.8918 \\
 IRN~\cite{xiao2020invertible} &90  & 33.38 / 0.9140 \\
 SAIN (Ours)&90 & \textbf{35.83} / \textbf{0.9410}\\\midrule
 Bicubic \& Bicubic &30 & 29.75 / 0.8244 \\
 Bicubic \& SRCNN~\cite{dong2015image} &30 & 28.47 / 0.8160\\
 Bicubic \& EDSR~\cite{lim2017enhanced} &30 & 29.62 / 0.8249\\
 Bicubic \& RDN~\cite{zhang2018residual} &30 & 29.64 / 0.8252\\
 Bicubic \& RCAN~\cite{zhang2018image} &30 & 29.54 / 0.8235\\
 CAR \& EDSR~\cite{sun2020learned}&30 & 28.03 / 0.7800\\
 IRN~\cite{xiao2020invertible} &30 &29.86 / 0.8303\\
SAIN (Ours)&30 & \textbf{33.15} / \textbf{0.9144}\\\bottomrule
\end{tabular}
    \caption{Quantitative results on DIV2K against WebP.}
    \label{tab:webpx2}
\end{table}

\section{Conclusion}
Existing image rescaling models are fragile to compression artifacts.
In this work, we present a novel self-asymmetric invertible network (SAIN) that is robust to lossy compression.
It approximates the downscaling and compression processes in one forward pass by virtue of the proposed E-InvBlock, and thus can inversely restore the compression distortions for improved upscaling.
We leverage a learnable GMM distribution to capture a generic knowledge shared across samples, and carefully design the loss functions to benefit approximations and restorations.
Extensive experiments prove that our model performs far better than previous methods under the distortion of standard image codecs and is flexible to be extended to other compression formats.

\bibliography{aaai23}

\twocolumn[{%
\renewcommand\twocolumn[1][]{#1}%
\begin{center}
    \centering
    \captionsetup{type=figure}
    \caption*{\LARGE \bf Supplementary Material}
    \includegraphics[width=\textwidth]{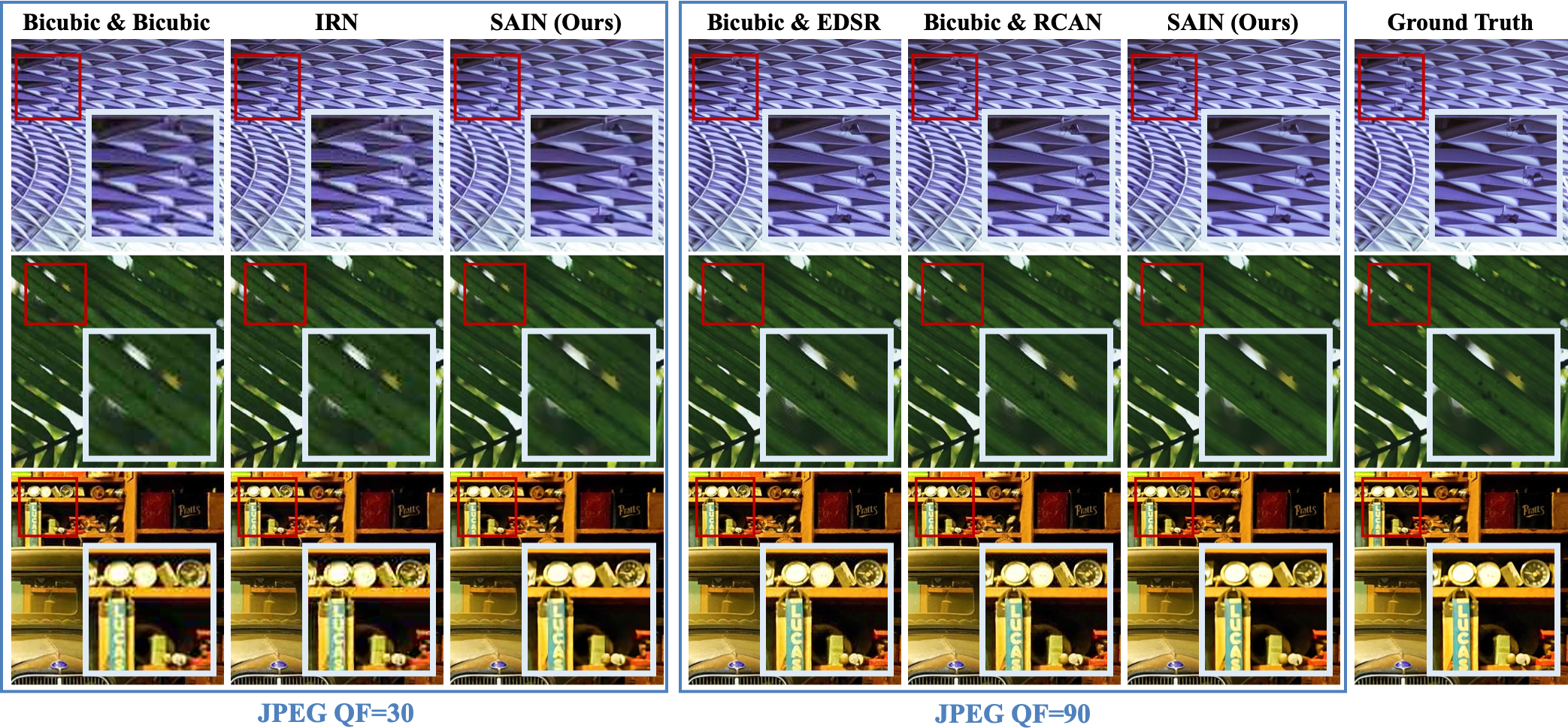}
    \captionof{figure}{Additional qualitative results of image rescaling ($\times2$) on DIV2K under distortion at different JPEG QFs.}\label{fig:jpegx2}
\end{center}%
\begin{center}
    \centering
    \captionsetup{type=figure}
    \includegraphics[width=\textwidth]{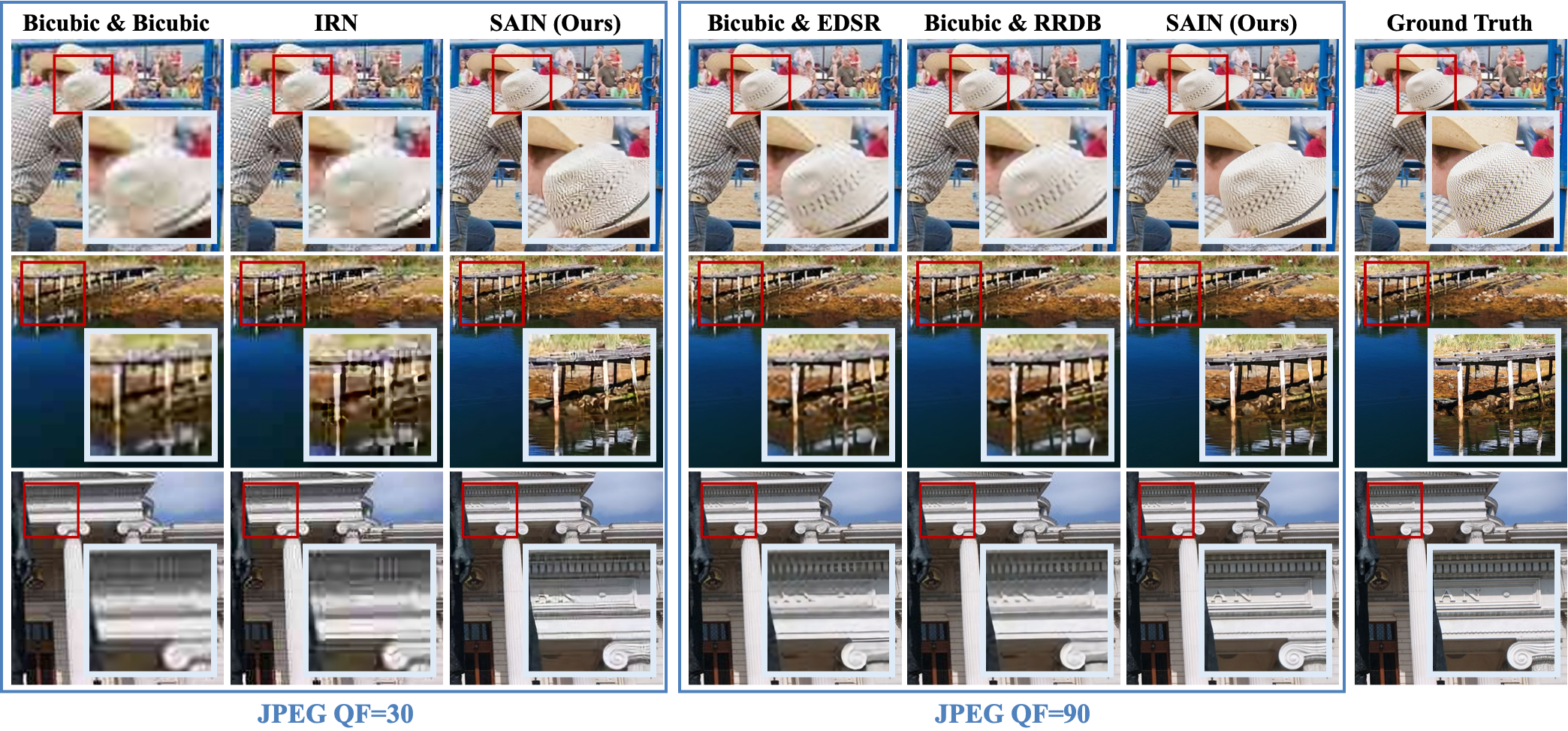}
    \captionof{figure}{Qualitative results of image rescaling ($\times4$) on DIV2K under distortion at different JPEG QFs.}\label{fig:jpegx4}
\end{center}%
}]

\appendix

\section{Log-Jacobian Determinant of E-InvBlock}
Flow-based methods usually requires a tractable Jacobian to directly maximize the log-likelihood of the input distribution.
Although we optimize our model in a different manner, the proposed E-InvBlock also allows for easy computation of the log-Jacobian determinant.

The forward step of the E-InvBlock consists of 3 steps:
\begin{equation}
y_{i}=x_{l}^{i}+\varphi\left(x_{h}^{i}\right)~;\\
\end{equation}
\begin{equation}
x_{l}^{i+1}=y_{i}-\phi\left(x_{h}^{i}\right)~;\\
\end{equation}
\begin{equation}
x_{h}^{i+1}=x_{h}^{i} \odot e^{\left(\rho\left(x_{l}^{i+1}\right)\right)}+\eta\left(x_{l}^{i+1}\right)~.
\end{equation}
And the corresponding Jacobians of each step are:
\begin{equation}
\small
J_{1}=\left[\begin{array}{ll}
1 & * \\
0 & 1
\end{array}\right] ; J_{2}=\left[\begin{array}{ll}
1 & * \\
0 & 1
\end{array}\right] ; J_{3}=\left[\begin{array}{ll}
1 & 0 \\
* & e^{\left(\rho\left(x_{l}^{i+1}\right)\right)}
\end{array}\right]~,
\end{equation}
where $*$ denotes the components whose exact forms are unimportant.
So the overall log-Jacobian determinant is:
\begin{equation}
\begin{aligned}
    &\log \left|\operatorname{det}\left(J_1*J_2*J_3\right) \right|\\
    =&\log \left|\operatorname{det}\left(J_1\right) \right|+\log \left|\operatorname{det}\left(J_2\right) \right|+\log \left|\operatorname{det}\left(J_3\right) \right|\\
    = &~0+0+\rho\left(x_{l}^{i+1}\right)~.
\end{aligned}
\end{equation}

\section{Visual Results under JPEG Distortion}

\paragraph{More Image Rescaling ($\times2$) Results}
We demonstrate additional visual results of the top methods in image rescaling ($\times2$) under JPEG distortions in Fig.\ref{fig:jpegx2}

\begin{figure}[t]
    \centering
    \includegraphics[width=\linewidth]{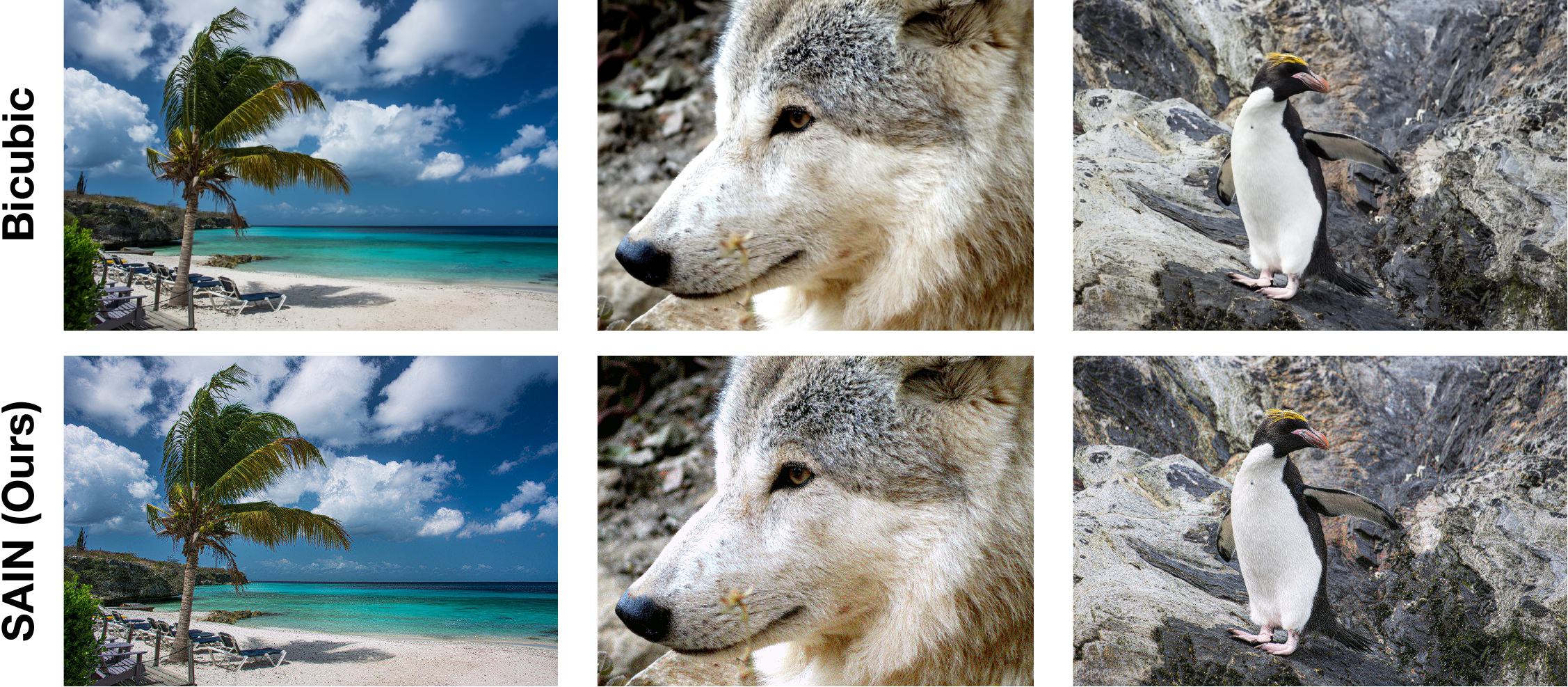}
    \caption{Visual results of downscaled images on DIV2K.}
    \label{fig:lr}
\end{figure}

\paragraph{Image Rescaling ($\times4$) Results}
Due to space limitation, the qualitative results of image rescaling ($\times4$) are omitted in the manuscript.
These results are shown in Fig.~\ref{fig:jpegx4}.
Apparently, at a larger scale like $\times4$, previous methods are even more fragile to the JPEG distortions.
In contrast, our model has the ability to obtain rather high-quality reconstructions.
\paragraph{Downscaled LR Results}
As shown in Fig.~\ref{fig:lr}, the downscaled results produced by our model is similar to the results of Bicubic downscaling in visual perception.

\section{Visual Results under WebP Distortion}
\begin{figure}[!t]
    \centering
    \includegraphics[width=\linewidth]{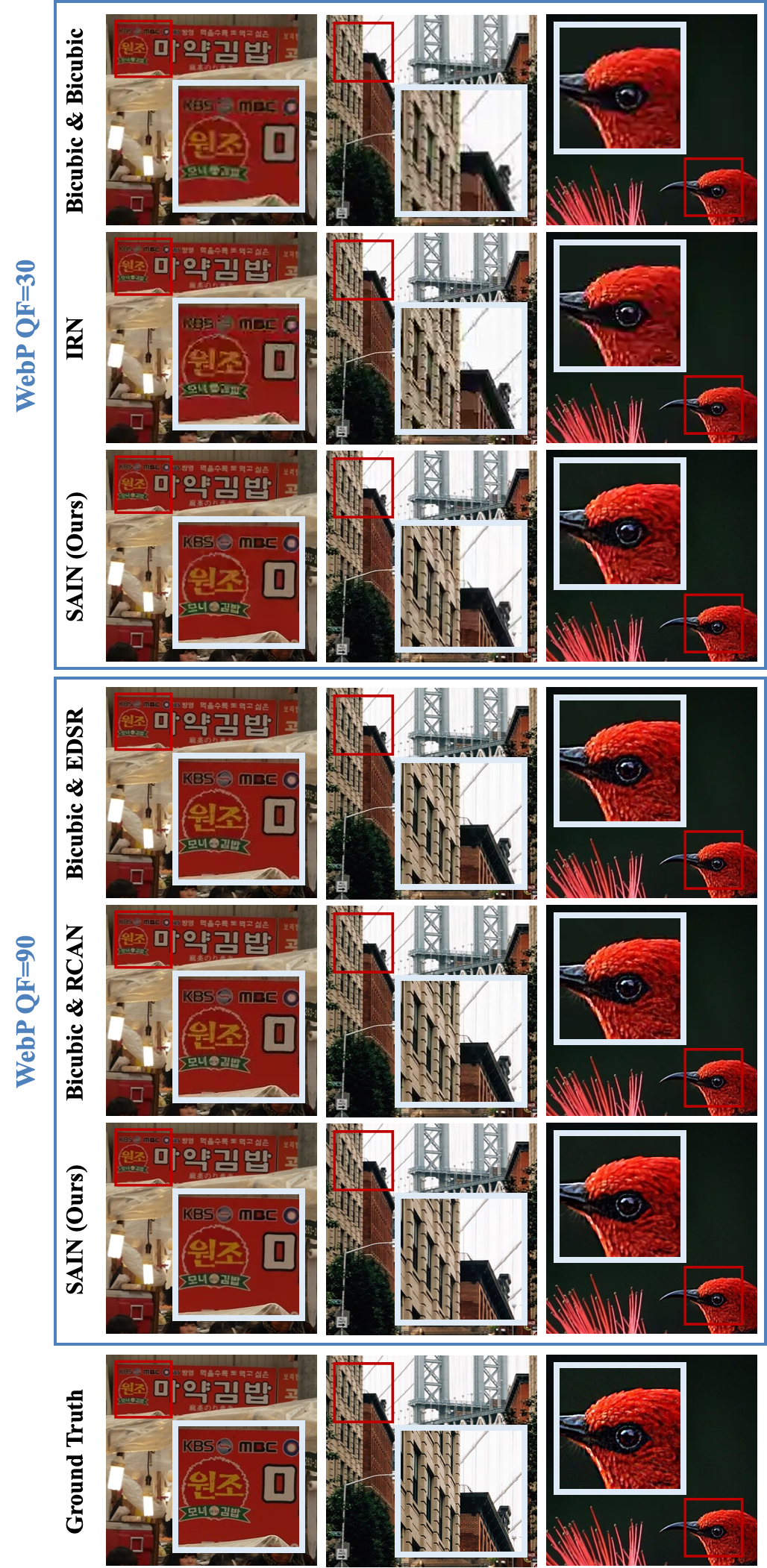}
    \caption{Qualitative results of image rescaling ($\times2$) on DIV2K under distortion at different WebP QFs.}
    \label{fig:webpx2}
\end{figure}

Since we incorporate real image compression to guide the training objectives, our method can also be easily applied to other compression formats.
Some visual demonstrations of image rescaling under WebP distortions are presented in Fig.~\ref{fig:webpx2}.
Although we leverage a differentiable JPEG implementation to surrogate the gradients of real image compression, our method still exhibits satisfactory robustness to the WebP compression.

\section{Effect of Training QFs}
In the main experiments, we empirically train our model with the compression QF fixed at 75.
As shown in Tab.~\ref{tab:qfs}, the model trained at QF=75 performs much better in higher QFs like 75 and 50.
We hypothesize that training with too low QF such as 25 severely degrades the quality of LR image representations and thus makes distortion recovery too difficult.
Given that the compression QF used in real scenarios is usually larger than 50, it is natural to focus more on the reconstruction performance at higher QFs. 

Besides, in our experiments, we also find that training with mixed compression QFs is also unfavorable to improve model performance. 

\begin{table}[t]
    \centering
    \begin{tabular}{l|ccccc}\toprule
    PSNR (dB)     & \small{(Testing QF=)} 75 & 50 & 25 \\\midrule
    \small{(Training QF=)} 75 & \textbf{35.10}& \textbf{33.17}& 30.89\\
    50 & 33.88 & 32.99 & \textbf{31.23}\\
    25 & 29.37 & 29.34 & 29.14 \\\bottomrule
    \end{tabular}
    \caption{Effect of different training QFs. 
    The results are evaluated on DIV2K under JPEG distortion.
    }
    \label{tab:qfs}
\end{table}

\end{document}